\documentclass{article}
\usepackage{ijcai22}

\usepackage{times}
\usepackage{soul}
\usepackage{url}
\usepackage[hidelinks]{hyperref}
\usepackage[utf8]{inputenc}
\usepackage[small]{caption}
\usepackage{graphicx}
\usepackage{amsmath}
\usepackage{amsthm}
\usepackage{booktabs}
\usepackage{algorithm}
\usepackage{algorithmic}
\title{Edge-Compatible Reinforcement Learning for Recommendations}

\author{
James E. Kostas$^1$
\and
Philip S. Thomas$^1$\And
Georgios Theocharous$^2$\\
\affiliations
$^1$College of Information and Computer Sciences, University of Massachusetts, Amherst, MA, USA\\
$^2$Adobe Research, San Jose, CA, USA\\
}
\usepackage{amsmath}
\usepackage{amsthm}
\usepackage{amssymb}
\usepackage{mathtools}
\usepackage{mathrsfs}
\mathtoolsset{showonlyrefs}

\usepackage{bm}

\usepackage{theoremref} % for \theoremref to nicely reference

\usepackage{courier} % For \texttt{foo} to put foo in Courier (for code / variables)

\usepackage{lipsum} % For dummy text

\usepackage{graphicx}
\usepackage{subcaption}
\usepackage[space]{grffile} % For spaces in image names

\usepackage[round]{natbib}

\usepackage{hyperref}

\usepackage{xcolor}
\definecolor{dark-red}{rgb}{0.4,0.15,0.15}
\definecolor{dark-blue}{rgb}{0,0,0.7}
\hypersetup{
    colorlinks, linkcolor={dark-blue},
    citecolor={dark-blue}, urlcolor={dark-blue}
}

\usepackage{siunitx} %commas in numbers
\newcommand{\numc}[1]{\num[group-separator={,}]{#1}}

\newcommand{\ith}{i^\text{th}}

\begin{document}

\maketitle

\begin{abstract}
    Most reinforcement learning (RL) recommendation systems designed for edge computing must either synchronize during recommendation selection or depend on an unprincipled patchwork collection of algorithms.
    In this work, we build on asynchronous coagent policy gradient algorithms \citep{kostas2020asynchronous} to propose a principled solution to this problem.
    The class of algorithms that we propose can be distributed over the internet and run asynchronously and in real-time.
    When a given edge fails to respond to a request for data with sufficient speed, this is not a problem; the algorithm is designed to function and learn in the edge setting, and network issues are part of this setting.
    The result is a principled, theoretically grounded RL algorithm designed to be distributed in and learn in this asynchronous environment.
    In this work, we describe this algorithm and a proposed class of architectures in detail, and demonstrate that they work well in practice in the asynchronous setting, even as the network quality degrades.
\end{abstract}

\section{Introduction}

An edge system typically consists of a central hub and geographically distributed edges.
The reason for having geographically distributed edges is to reduce response times between users and a web-service such as a reinforcement learning (RL) recommendation system.
The benefit of faster response times comes at a cost, both for training and execution: one can no longer deploy centralized machine learning algorithms.
If one naively trains independent models using local edge data, suboptimal solutions will be computed.
Even if one managed to train a consistent global model, which could be updated on a regular basis (for example, every night), execution would still be a problem, since network latencies vary and temporary outages (for example, a server going offline) can occur.
In real time when the algorithm needs to choose an action, it may need to consider dynamic or historical features of assets from all the edges.
For example, to recommend an item at the current edge the algorithm may need to know recent historical performance of the item from other edges.
Another possibility is that all documents may not reside on all edges; for example, to recommend a document at the current edge, the algorithm may need to send a signal to an algorithm component residing on another edge, and then receive relevant document information from that remote component.

This work proposes the use of asynchronous coagent networks \citep{kostas2020asynchronous}, a type of stochastic neural network for reinforcement learning (RL), to provide recommendations while the network is distributed over the internet.
These networks provide principled, theoretically-ground learning rules for training in the distributed, edge-compatible, asynchronous recommender systems setting, where outages and variable or extreme latencies may occur.

The three primary contributions of this paper are:
\textbf{1)} Previous work only deals with asynchronous networks from a theoretical perspective, and does not show that they work well in practice.
This work shows that these networks work in practice for interesting, high-dimensional problems based on real-world data.
\textbf{2)} Earlier work does not suggest the possibility of distributing asynchronous coagent networks over the internet or computer networks, so the application of these asynchronous networks to edge computing is another contribution.
In addition to proposing this new use case, we demonstrate empirically that this class of algorithms works well in practice for this use case.
\textbf{3)} Another contribution is the class of architectures we propose for the (asynchronous) edge recommender system setting.

\section{Related Work}

The use of machine learning to optimize recommender systems without expert human supervision dates back at least to the work of \citet{joachims2002optimizing}, who proposed a support vector machine algorithm to optimize clickthrough rates for search engine results.
%
% \citet{craswell2008experimental} study the positioning of recommendations on a results page (specifically, they study the positioning of results for search engines).  They consider several hypotheses, and propose that a \emph{cascade model}, ``where users view results from top to bottom and leave as soon as they see a worthwhile document'', best explains the real-world data.

Many works have studied the RL recommendation problem.
\citet{choi2018reinforcement} use biclustering to reduce the dimensionalities of the state and action spaces, thus making the problem more tractable for RL algorithms.
\citet{chen2019top} propose an RL-based recommender system that scales to a large action space and that corrects for bias induced by learning from data collected based on recommendations from an earlier recommender system.
\citet{theocharous2009tractable} study a partially-observable setting in which an algorithm must teach human students a task. This environment is modeled as a \emph{partially-observable Markov decision process} (POMDP), and the students have an observable internal state (for example, confusion or boredom), which affects the transition function.
They propose learning expert policies for different settings, and a \emph{switching} policy that learns when to switch between these experts.
This approach bears a deep resemblance to the coagent networks used in this paper.
\citet{SlateQ} also study the RL recommendation problem, with a particular focus on the setting in which an algorithm should consider the long-term effects of its actions.
They assume that they are given the user-choice model (the model that describes how a given user will interact with a set of recommendations), and propose an algorithm, SlateQ, which makes the problem tractible under certain assumptions.
\citet{zhao2018deep} study the recommender systems setting in which users can provide real-time feedback.
They propose an actor-critic-like learning algorithm and deep recurrent architecture to solve the problem.
\citet{theocharous2018scalar} develop a posterior sampling for reinforcement learning algorithm that is effective at learning to give personalized recommendations.  They prove a regret bound and empirically show that the algorithm is effective.
\citet{theocharous2020reinforcement} study various aspects of the RL recommendation problem, including modeling user behavior, offline high confidence evaluation, constraint optimization, and non-stationarity.

Other works study bandit problems (settings where the myopic policy is optimal) and ranking problems.
\citet{swaminathan2017off} study the recommendation problem in the contextual bandit setting.
They propose the pseudoinverse estimator, which estimates the performance of a policy, thus providing a method for evaluating a policy using off-policy data.
They prove that their estimator is unbiased under certain conditions, and they show that it performs well empirically.
\citet{ai2018learning} propose a recurrent architecture and algorithm to refine and re-rank the results from another ranking algorithm.
\citet{bello2018seq2slate} study the ranking problem, and formalize it as a sequence prediction problem.  They incorporate pointer networks, which are a model specialized for the ranking problem, into an algorithm designed to solve the problem.
They show that their algorithm is empirically effective in several settings.
\citet{jiang2018beyond} point out that many popular recommender algorithms employ greedy ranking; these algorithms cannot account for biases in a slate layout and in interactions between recommendations (for example,  the effect of the contrast between two recommendations, or the ``relative attractiventess'' of two recommendations).
They propose an algorithm that uses  conditional generative modeling to directly generate an entire slate of recommendations, so as to correctly account for these biases and interactions.
\citet{rhuggenaath2020algorithms} study the slate bandit setting in which rewards are not a simple function of the individual components of the slate.  In other words, the authors eliminate common assumptions about the reward which make the combinatorial slate problem more feasible.  They make an independence assumption about the structure of the reward function, and use this assumption to propose an algorithm with sub-linear regret with respect to the time horizon.
%
% To accomplish this, they require that the ordering of display positions is fixed throughout training and inference times.
%
% \citet{xie2017investigating} employ eye-tracking to study how users examine image search results.
% %
% Based on their results, they conclude that users have several previously-unknown biases when examining image search results.
% %
% They propose several methods of optimizing image search based on their findings.

% \citet{xie2018constructing} conduct a user study for image search, and propose a grid-based user browsing model (GUBM) for recommender systems that return 2-D (``grid'') results, based on user interaction signals, such as click and hover.

However, none of these works address the recommender setting which we study: the setting in which the algorithm and/or data itself must be distributed and run in an asynchronous manner.

\emph{Federated learning} \citep{qi2021federated} is distributed RL, but in an entirely different sense of the word ``distributed'' than in this paper; Section \ref{sec:federated_learning} of the supplementary material discusses this distinction in more detail.

% \citet{kostas2020asynchronous} give theoretically-grounded learning rules for training asynchronous networks for RL; we leverage these learning rules for our approach.

\section{Background}

We study the setting in which the recommendation problem is a \emph{Markov decision process} (MDP).
Some problems in our setting are \emph{contextual bandits}, which are MDPs with only one timestep per episode.
We denote the state and action spaces as $\mathcal S$ and $\mathcal A$ respectively.
We denote the timestep as $t$.
The random variables $S_t \in \mathcal S$, $A_t \in \mathcal A$, and $R_t \in \mathbb R$ denote the state, action, and reward at time $t$, respectively.
The transition function, $P:\mathcal S \times \mathcal A \times \mathcal S \to \mathbb R$, gives the probability of transition to a state, given a state and action: $P(s, a, s') \coloneqq \Pr(S_{t+1} = s' | S_t = s, A_t = a)$.
We write the reward discount parameter as $\gamma \in [0, 1]$.
The agent is parameterized by $\theta \in \Theta$, where $\Theta$ is the feasible set.
The agent's goal is to maximize the objective $J: \Theta \to \mathbb R$, which is defined as $J(\theta) \coloneqq \mathbb E [\sum_{t=0}^\infty \gamma^t R_t | \theta]$; where ``given $\theta$'' means that an agent parameterized by $\theta$ chooses the actions.
\emph{Policy gradient algorithms} are a popular class of RL algorithms that aim to maximize the objective by performing stochastic gradient ascent using estimates of the objective's derivative, $\nabla J(\theta)$.

\subsection{Asynchronous Coagent Networks}

We propose a unified distributed training and execution approach based on asynchronous coagent networks \citep{kostas2020asynchronous}.  Coagent networks are a type of asynchronous stochastic neural network that are comprised of conjugate agents, or coagents, operating asynchronously in \emph{continuous time}.
Each coagent is an RL algorithm learning and acting cooperatively with the other coagents in its network.
During a coagent's \emph{execution}, it takes as input the state of the environment and the outputs from the other coagents (not necessarily the whole state space or the outputs of all other coagents, it could take a small subset of the union of these spaces), and computes some output, which the coagent continuously outputs until its next execution.
The output of the coagent is fed into other coagents and/or the network's output (i.e., the network's action).
A coagent's probability (mass and/or density) of updating at any time can be a deterministic or stochastic function of its input and the state of the environment.
For example, one coagent might be designed to execute at $10$ Hz, another coagent might be designed to execute each microsecond with a probability of $10^{-7}$, and a third might execute at times based on some (stochastic or deterministic) function of the environment's state and/or the outputs of the coagents.

During training, each coagent computes a \emph{local gradient}; the gradient rules are straightforward to derive \citep{kostas2020asynchronous}.  The asynchronous coagent policy gradient theorem states that if each coagent updates its parameters using its local gradient, then the entire network will be updated as if the global policy gradient was computed. The theory holds for asynchronous networks where the units in the neural network do not execute simultaneously or at the same rate.

Additional intuition about this theorem (which enables the class of asynchronous recommender algorithms that we propose) follows.
First, consider the complexity of the environment that each coagent faces:
    \textbf{1)} Each coagent is embedded in an asynchronous \emph{continuous-time} network of coagents; \textbf{2)} those other coagents are learning and updating as time goes on, which makes the environment around each coagent \emph{non-stationary};
%(in other words, each coagent's environment, which consists of the combination of the other coagents and the original MDP environment, is non-stationary);
and \textbf{3)} the network as a whole is interacting with an environment (i.e., the original environment of the MDP).
The theorem states that if each coagent ignores the complexity of this setup and simply runs an unbiased policy gradient algorithm (as if it were in a stationary discrete-time MDP), treating its inputs as the state space, its outputs as the action space, and the (discounted sum of) rewards between the coagent's executions as the rewards between timesteps, then the network as a whole will follow an unbiased policy gradient.
See the work of \citet{kostas2020asynchronous} for a more formal description.

See Section \ref{sec:example_algo} for an example of this class of algorithm.
The asynchronous nature of training and execution makes the approach natural for distributed implementations and particularly for our edge setting.

\section{Approach}
\label{sec:approach}

This section describes the architecture we propose as well as an example asynchronous coagent policy gradient algorithm.

\subsection{Architecture}

We propose an architecture in which every edge has its own coagents that compute recommendations scores for each of its items.
Parameters are shared between the corresponding coagents for each document.
Each edge has a coagent that decides how to combine the computations of adjacent edges.

The learning algorithm accounts for all network delays in a theoretically-grounded way.
For example, it will account for the situation where it had to give a suboptimal recommendation due to a delay and how that delay and suboptimal recommendation should influence the follow-up recommendation:
The learning rule and corresponding policy gradient naturally account for this situation and correctly provide an unbiased gradient.
In other words, the learning rule simply views these limitations (e.g., delays and any resulting suboptimal recommendations) as part of the objective (via the asynchronous learning rule), and optimizes accordingly.

\begin{figure}
    \includegraphics[width=0.6\linewidth]{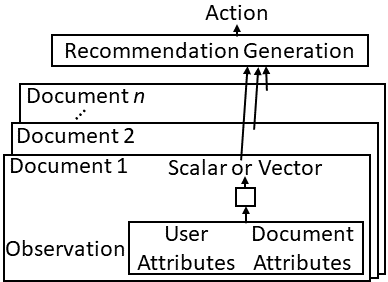}
    \centering
\caption{A high-level view of our recommendation architecture.  The small empty square represents a coagent or group of coagents. At a given time-step, for each document, the network takes user attributes (including the query, if applicable) and document attributes as inputs.  These observations are processed, and a scalar or vector is computed for the document.
The parameters used to generate each document's scalar or vector are the same (that is, the parameters are \emph{shared} across the network).
The recommendation generation component takes these scalars as input (one per document) and outputs a recommendation.  The recommendation generation process may be a simple fixed algorithm: a sorting of scalars (one per document) may be effective if the user choice model can be assumed to be a cascade \citep{craswell2008experimental} or logit model \citep{SlateQ}.  A softmax could be used (similar to the approach of \citet{chen2019top}) for more efficient exploration.
%
% Alternatively, in the case of slate recommendations, the slate generation component could be parameterized and learned over.  For example, one could parameterize this component to be a softmax and sort, and hold those parameters constant during the initial part of the learning process.  Then, after the rest of the network reaches a performance threshold, those parameters could be unlocked and fine-tuned with the learning algorithm so that the network can achieve better performance and account for effects such as positional biases or interference between documents in a 2-D slate.
}
\label{fig:architecture}
\end{figure}

\begin{figure}
    \includegraphics[width=1.0\linewidth]{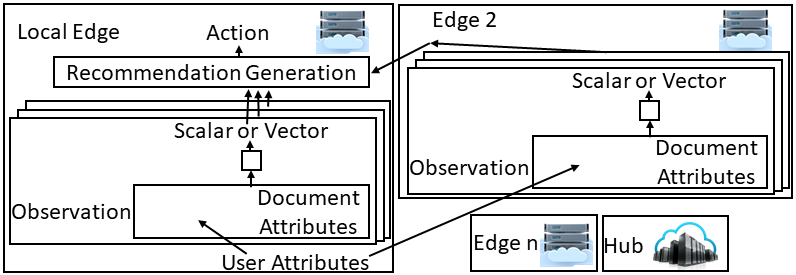}
    \centering
\caption{Each user/document computation may take place on a different edge, and the final computation which computes the recommendation takes the results and computes the action (recommendation) on the local edge.
Connections for the local edge and edge $2$ are visualized as arrows; connections for edge $n$ and the hub are not visualized, but the process is the same as for edge $2$.
Since some results may not arrive until after the recommendations must be displayed to the user (in other words, at any given time, the algorithm may not be able to communicate completely with itself and/or may not be able to access some documents), the algorithm must learn to achieve the best result it can for each recommendation, given the limitations of the edge setting.}
\label{fig:async_architecture}
\end{figure}

Figure \ref{fig:architecture} shows a high-level view of the architecture we propose.
Figure \ref{fig:async_architecture} demonstrates how this architecture can be distributed in the edge setting; see the captions for more details.

\subsection{Example Algorithm in Continuous Time}
\label{sec:example_algo}

In this setting, time is continuous: no global syncing of the algorithm is ever necessary for action selection.  Each component of the network (coagent) simply executes (computes its action) whenever a new signal reaches it.

The following update describes a simple example asynchronous coagent learning algorithm based on REINFORCE \citep{williams1992simple}.
Let $\Theta_i$ be the feasible set for coagent $i$, $\theta_i \in \Theta_i$ be the parameters of coagent $i$, $\alpha$ be the step size, and $G_t$ be the return (that is, the cumulative discounted reward%, analogous to $\sum_{t=0}^\infty \gamma^t R_t$
) up to time $t \in \mathbb R$ (notice that because time is continuous, $t$ does not take a value in the integers, as is typical in RL, but instead takes a value in the reals). Let $\mathcal U_i$ be the output space of the $\ith$ coagent, and let $\mathcal X_i$ be the space of inputs to the $\ith$ coagent (a subset of $\mathcal S \times \mathcal U_1 \times \mathcal U_2 \times \cdots \times U_m$, where $m$ is the number of coagents).
Let the random variable $X^i_t \in \mathcal X_i$ be the inputs to the coagent at time $t$, and the random variable $U^i_t \in \mathcal U_i$ be the action (output) of the coagent at time $t$.  Let $n$ denote the number of times a coagent executes during a sequence of interactions with the environment.
Let $\pi_i:\mathcal X_i \times \mathcal U_i \times \Theta_i \to \mathbb R$ be the policy of the $\ith$ coagent, which gives the probability of an output $u \in \mathcal U_i$ given an input in $x \in \mathcal X_i$ and a set of parameters in $\theta_i \in \Theta_i$: $\pi_i(x, u, \theta_i) \coloneqq \Pr(U^i_t = u | X^i_t = x, \theta_i)$.
Let $t_1, t_2, \dotsc, t_n$ be the times of the coagent's first execution, second execution, $\dotsc$, and $n^\text{th}$ execution.  The update each coagent will follow after a sequence of environment interactions is:

\begin{align}
    \theta_i \gets \theta_i + \alpha \!\!\!\!\!\!\!\!\!\!\! \sum_{t \in \{t_1, t_2, \dotsc, t_n\}} \!\!\!\!\!\!\!\!\!\!\! \gamma^t G_t \left(\frac{\partial \ln\left ( \pi_i\left ( X^i_t, U^i_t, \theta_i \right ) \right )}{\partial \theta_i}\right). 
\end{align}

In practice, a more sophisticated policy gradient algorithm than this simple variant of REINFORCE may be used; see supplementary material Section \ref{sec:algo_details} for further discussion.

% \subsection{Other Approaches}

% One existing solution is to simply use a synchronous algorithm; most existing recommender systems use this approach. Some downsides of this approach include the inability to leverage edge computing for faster response time and distributed storage and compute.  Additionally, if the documents are stored in a distributed manner, then the synchronous algorithm must wait for all edges to reply, reducing the responsiveness of the system.

% One class of distributed solutions is to simply use a ``bag-of-algorithms" approach. In other words, one can patch together several learning algorithms and distribute the resulting network. However, these kinds of solutions are not theoretically grounded, and so may diverge or exhibit undesirable properties when deployed. We are not aware of any unified and theoretically-grounded distributed solution to the problem other than our own.

\section{Results}

\newcommand{\subfigwidth}{.95}

\begin{figure*}
    \begin{subfigure}{.49\textwidth}
    \centering
    \includegraphics[width=\subfigwidth\linewidth]{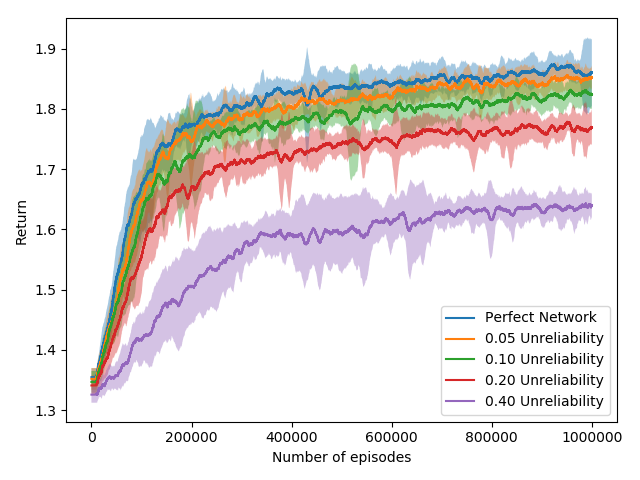}
    \caption{MSLR-WEB10K bandit results}
    \end{subfigure}
    \begin{subfigure}{.49\textwidth}
    \centering
    \includegraphics[width=\subfigwidth\linewidth]{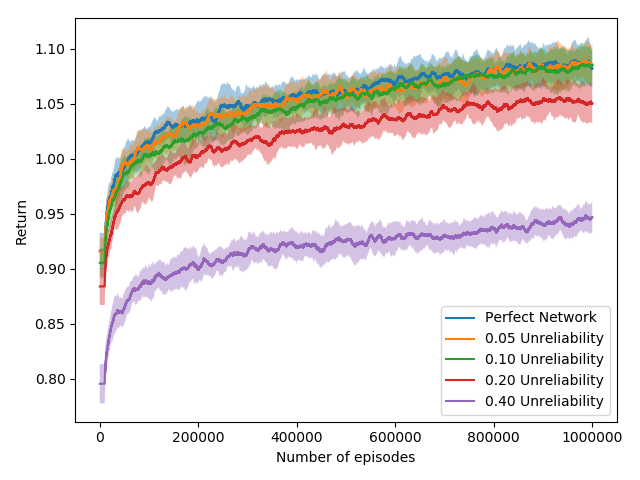}
    \caption{MQ2008 bandit results}
    \end{subfigure}
\caption{Asynchronous edge setting results.  These demonstrate that the algorithm can continue to learn effective recommendation strategies in an increasingly asynchronous edge setting. Note that these results do \emph{not} indicate that the algorithm is performing worse given more unreliability.  Rather, as the unreliability increases, the environment is becoming more challenging, degrading the agent's ability to access documents and to communicate with itself; even an optimal policy would have progressively lower return as unreliability increases.  These results demonstrate that our algorithm can learn effective recommendation strategies even in harsh, extremely asynchronous conditions.}
\label{fig:async_results}
\end{figure*}

In this section, we give results demonstrating that the algorithm and architecture proposed learn effectively in the asynchronous edge setting.
Note that these results are \emph{not} necessarily intended to demonstrate state-of-the-art results for the standard RL recommender-system setting. 
Instead, these results show that the proposed algorithm can learn and function effectively in the asynchronous setting.
See supplementary material Section \ref{sec:algo_details} for algorithm and architecture details not covered in Section \ref{sec:approach}.

Since no other algorithm we are aware of is designed to handle our asynchronous setting, there is no obvious baseline to compare against.
%
% Furthermore, the comparison to a baseline is not an essential part of the paper, since the results are not intended to demonstrate state-of-the-art performance for the standard RL recommender-system setting.
%
However, for completeness, we propose a naive baseline to compare our algorithm to, and give the results in Section \ref{sec:baseline} (and further results in supplementary material Section \ref{sec:baseline_mslr}).

All experiments were conducted using a set of simulators developed from the MSLR-WEB10K and the MQ2008 datasets \citep{datasets_2013}.
See Section \ref{sec:simulator} of the supplementary material for simulator details.

All learning curves are plotted from 30 runs (that is, 30 trials), and all error bars show the standard deviation (between runs).
Because of the stochastic nature of recommender systems and the resulting highly stochastic returns, for readability, each plotted data point is the running average of the previous $\numc{10000}$ episodes (or, in the case of the first $\numc{10000}$ episodes, each data point is simply the average of the first $\numc{10000}$ episodes so as to avoid excessive noise at the beginning of each plot).

\subsection{Asynchronous Bandit Simulations}
\label{sec:async_results}

\begin{figure*}
    \begin{subfigure}{.49\textwidth}
    \centering
    \includegraphics[width=\subfigwidth\linewidth]{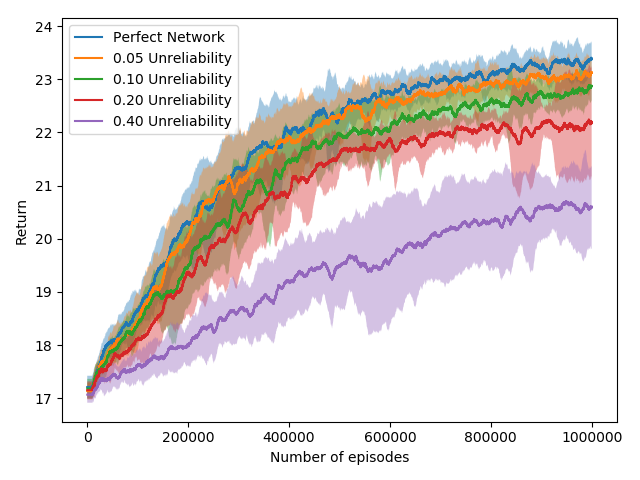}
    \caption{MSLR-WEB10K RL results}
    \end{subfigure}
    \begin{subfigure}{.49\textwidth}
    \centering
    \includegraphics[width=\subfigwidth\linewidth]{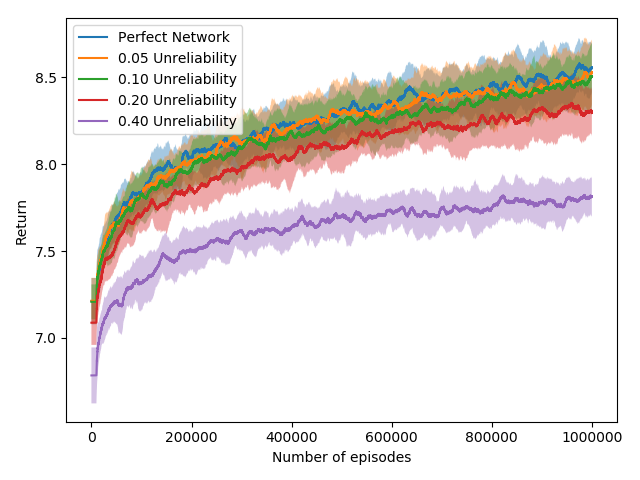}
    \caption{MQ2008 RL results}
    \end{subfigure}
\caption{RL results.  Similar to the results above, these results demonstrate that the algorithm can learn effective recommendation strategies in an increasingly asynchronous edge setting.}
\label{fig:RL_results}
\end{figure*}

In this section, we demonstrate that the algorithm can function in the asynchronous edge setting, as described above.
We created different versions of the simulators based on an \emph{unreliability} parameter.  An unreliability of $0.0$ means that the network is fully synchronous, and an unreliability of $1.0$ means that the network is fully asynchronous (which would make communication between edges, and thus recommendations, impossible).
More generally, an unreliability of $p \in [0, 1]$ means that, with probability $p$, each document/query pair will not compute a scalar in time to respond to the local edge before it must display results to the user (and so that document will be unavailable to the algorithm for that timestep).
The results are shown in Figure \ref{fig:async_results}; they demonstrate that the algorithm can continue to learn effective recommendation strategies in an increasingly harsh asynchronous edge setting.

\subsection{Asynchronous RL Simulations}

Next, we added a temporal aspect to the simulation; see Sections \ref{sec:rl_simulation_mslr_details} and \ref{sec:rl_simulation_letor_details} of the supplemental material for details.
The results are displayed in Figure \ref{fig:RL_results}.  These results demonstrate that the algorithm learns effective recommendation strategies in this setting even as network quality degrades.
As in Section \ref{sec:async_results}, note that the results do not indicate that the algorithm is getting worse in more asynchronous settings.
Rather, the problem and environment are becoming more difficult as the problem becomes more asynchronous (such that even an optimal policy would result in lower returns), and the algorithm is learning effective recommendation strategies despite this fact.

\begin{figure}
    \centering
    \includegraphics[width=1.0\linewidth]{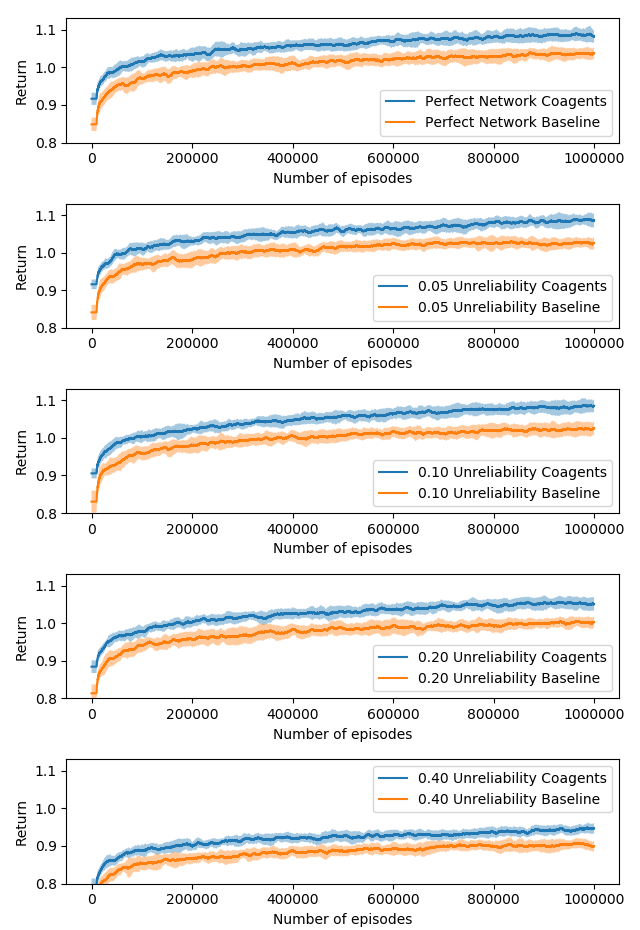}
    \caption{Baseline comparison for the MQ2008 bandit simulator.  The axes and scales on all five subplots are the same for ease of comparison (as are the axes and scales within the other baseline plots in Figures \ref{fig:baseline_letor_non_bandit}, \ref{fig:baseline_mslr_bandit}, and \ref{fig:baseline_mslr_non_bandit}).}
    \label{fig:baseline_letor_bandit}
\end{figure}

\begin{figure}
    \centering
    \includegraphics[width=1.0\linewidth]{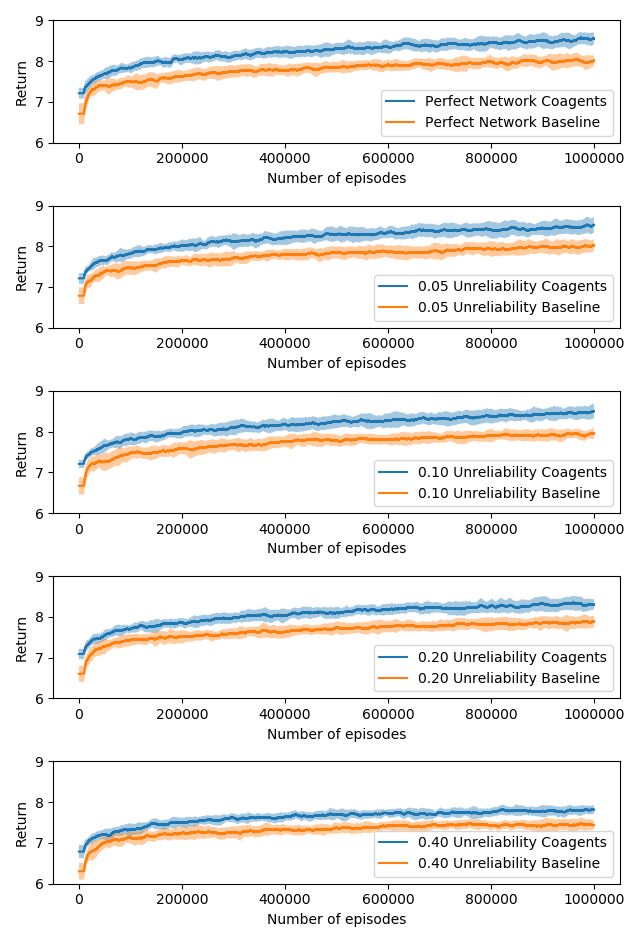}
    \caption{Baseline comparison for the MQ2008 RL simulator}
    \label{fig:baseline_letor_non_bandit}
\end{figure}

\begin{figure}
    \centering
    \includegraphics[width=1.0\linewidth]{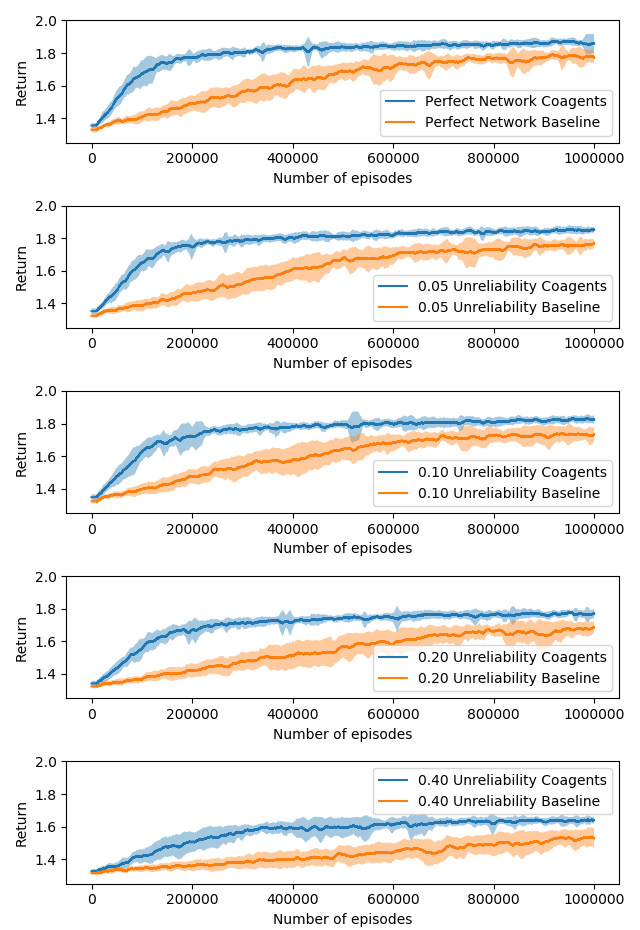}
    \caption{Baseline comparison for the MSLR-Web10K bandit simulator}
    \label{fig:baseline_mslr_bandit}
\end{figure}

\subsection{Baseline Comparison}
\label{sec:baseline}

Since no other algorithm we are aware of is designed for our asynchronous setting, there is no obvious baseline to use for comparison.
One naive but principled way of approaching the problem is to perform coordinate descent: fix the parameters of all but one edge, and then optimize the parameters of that one edge.
Then, repeat this process by looping over the edges, switching which one is being optimized each episode.

In the parameter-sharing case, which we use for the baseline to make the comparison to our algorithm as fair as possible, these optimizations affect the parameters of other edges, since the parameters are shared across edges.
Nonetheless, unlike for a coagent learning rule, each optimization step is with respect to a \emph{single} edge, and ignores the inputs and outputs of the other edges for the purposes of the parameter update.

To make the comparison as fair as possible, the coordinate descent baseline uses the exact same architecture (including parameter sharing), RL learning algorithm (REINFORCE, except that it is performing coordinate descent rather than a coagent update), and policy.

Figures \ref{fig:baseline_letor_bandit}, \ref{fig:baseline_letor_non_bandit}, and  \ref{fig:baseline_mslr_bandit} show the results for the two MQ2008 simulators and the MSLR-WEB10K bandit simulator (because of space limitations, the MSLR-WEB10K RL simulator results are given in supplementary material Section \ref{sec:baseline_mslr}).
These results show that the architecture (which both the baseline and our algorithm share) is effective at solving these problems.
The results also show that the asynchronous coagent-based algorithm substantially outperforms the naive baseline in all four simulators, for all settings of network unreliability.

\section{Future Work}

These algorithms bridge the gap between local and global RL recommender algorithms: they are neither a purely local algorithm (for example, running and learning independently on one server) nor are they a purely global algorithm (for example, a distributed policy that must fully sync up across servers and the edge to make a recommendation).
Instead, this class of algorithms is a principled blend of these extremes: these algorithms are able to learn to recommend during suboptimal network conditions, but are also able to learn to leverage the distributed data and/or the compute of other servers and edges if possible, using a theoretically-grounded approach that fully accounts for the complexity of this asynchronous real-time setting.

Our approach could enable the creation of algorithms to solve recommender problems where the asynchronous real-time nature is not only a function of the server and edge network, but also a function of the recommendation interface, the user, and the objective.
Consider a recommender problem where the agent's recommendations can be updated and refined in real-time.
For example, perhaps a near-optimal policy would first rapidly make a ``best-guess'' recommendation constrained by compute and/or network limitations (so that the user is not left without results while more compute occurs or data is retrieved).
A few seconds later, after data is retrieved from distant servers and/or a more sophisticated computation has been run, if the user has not yet selected an option, the policy may choose to refine or update this recommendation.
Recommender algorithms based on asynchronous coagent networks are well-suited to this type of real-time setting and offer a principled approach to this challenging class of problems.

\section{Summary and Conclusions}

In this paper, we propose a distributed class of RL recommender algorithms and architectures based on asynchronous coagent networks.
This class of algorithms is designed to learn and function in an asynchronous edge recommender setting using principled and theoretically-grounded learning rules.
Using simulations based on real-world data, we show that this approach works well in practice, even when the asynchronous edge setting interferes with the ability of the algorithm to access documents and to communicate with itself.

\bibliographystyle{unsrtnat}
\bibliography{references}

\newpage
% \,
% \newpage
\appendix

\section{Federated Learning}
\label{sec:federated_learning}

Federated learning \citep{qi2021federated} is distributed RL, but in an entirely different sense of the word ``distributed'' than in this paper.
Federated learning algorithms allow for the distribution of the training of a model among different parties, typically for the purpose of keeping the training data private.
Typically, for federated learning algorithms, copies of a model are distributed to multiple parties to be run on different (private) data, and the resulting model update (e.g., a gradient) is then computed based on the information the different parties send back.
Most federated learning algorithms are in fact \emph{synchronous} in the terminology of this paper, since action selection must occur synchronously.
Unlike asynchronous coagent networks, federated learning does not aim to distribute a single instance of a model over a network or run asynchronously for the purpose of computing actions.

So while the distributed nature of federated learning algorithms might sound superficially similar to asynchronous coagent networks, in fact the meanings of and motivations behind the ``distributed'' natures of these two classes of algorithms are entirely orthogonal and unrelated.

\section{Simulator Details}
\label{sec:simulator}

\subsection{MSLR-WEB10K Simulator Details}
\label{sec:simulator_mslr_bandit_details}

This simulator is based on approximately $\numc{235000}$ query-url pairs drawn from the MSLR-WEB10K dataset \citep{datasets_2013}.
The dataset provides \emph{relevance} scores for each query-url pair; these scores take a value in $\{0, 1, 2, 3, 4\}$, where $0$ is the least relevant and $4$ is the most relevant.
We use these scores as rewards; for a given query, when the agent selects a given document, the corresponding relevance (in $\{0, 1, 2, 3, 4\}$) is the reward.
All features are normalized to the range $[-1, 1]$.

For each timestep (which corresponds to a whole episode for the bandit setting), the simulator randomly selects a query.
The simulator then presents the agent with five possible documents to recommend (prepossessing to narrow down the options for the agent is a well-established technique, see, for example, the work of \citet{SlateQ}).
To make the problem interesting, and to incentivize non-myopic behavior (see the RL simulator variant below), the simulator tries to select documents with a large range of relevance.
Specifically, it selects documents with the following priority: $4, 0, 2, 3, 1$.
So, if documents with all relevance scores exist for the query, then the five documents will have the five different possible relevance scores.
In the case where the simulator does not contain a document with a given relevance for the query, the simulator loops back to the beginning of the priority list above to find replacement document(s).
For example, if a document with relevance $2$ cannot be found, an extra document with relevance $4$ will be provided, since $4$ is at the beginning of the priority list above.
Another example: if documents cannot be found with relevance $4$ or $2$, then the simulator will provide an extra document with relevance $0$ and an extra document with relevance $3$ (since those are the two highest priority relevance scores after $4$ and $2$).

\subsection{MSLR-WEB10K RL Simulator Details}
\label{sec:rl_simulation_mslr_details}

For the RL variety of the simulator, we added a temporal aspect to the simulation, inspired by the problem described by \citet{SlateQ}.
In this setting, a dimension is added to the state space which represents some aspect of the user's internal state; below, we refer to this as the user state (USE).
Similar to the experiments described by \citet{SlateQ}, recommending a low relevance document (that is, a non-myopic document) stochastically increases the USE over the course of an episode.
Specifically, the USE starts each episode with a value of $0$, and, on timestep $t$, stochastically increases by a random variable $B_t$ when a document with relevance $0$ or $1$ is recommended, where $B_t$ is sampled from $U(\{0.0, 0.4, 0.8\})$, where $U$ denotes the discrete uniform distribution.
If the USE is greater than or equal to $0.8$, the reward given for each timestep is multiplied by ten for the duration of the episode.
Episodes last for five user-system interactions.
The result is that a less myopic policy that focuses on increasing the USE in addition to making good myopic recommendations will be more effective than even the optimal myopic policy.
As in the work by \citet{SlateQ}, the increase of the USE might represent nudging the user's interests towards higher-value types of documents.
Alternatively, the increase of the USE might represent increased interest or trust in the recommender system, the advantages of showing more diverse recommendations, or other similar benefits that can be gained from making non-myopic recommendations.

\subsection{MQ2008 Simulator Details}
\label{sec:simulator_letor_bandit_details}

This simulator is based on the MQ2008 dataset \citep{datasets_2013}.
In this dataset, the possible relevance scores (and thus simulator rewards) are $\{0, 1, 2\}$.
On each timestep, the simulator selects documents with the following priority: $0, 2, 1$.
All other details are the same as those described in Section \ref{sec:simulator_mslr_bandit_details}.

\subsection{MQ2008 RL Simulator Details}
\label{sec:rl_simulation_letor_details}

This simulator modifies the MQ2008 simulator in the same way that the MSLR-WEB10 RL simulator modifies the (bandit) MSLR-WEB10 simulator.
The only differences are that: 1) when the USE is greater than or equal to $0.8$, the reward is multiplied by five (instead of ten), and 2) only documents with a relevance score of 0 may modify the USE (instead of relevance scores of 0 and 1).
%
% As for the MSLR-WEB10K problem, this choice was made to balance the problem (specifically, so that the random policy would result in a considerably lower average return than the optimal myopic policy, while still maintaining a considerable advantage for policies that maximize the USE).

\section{Algorithm and Architecture Details}
\label{sec:algo_details}

All coagent algorithms use a learning rule based on REINFORCE \citep{williams1992simple} (see Section \ref{sec:example_algo}).
While more sophisticated, lower-variance learning algorithms may be used in practice, the simple asynchronous coagent version of the REINFORCE learning rule is sufficient for the purposes of this work (to illustrate the fact that these algorithms can be distributed and run asynchronously, even in poor network conditions).
The learning rule is also principled in that it is unbiased (unlike most more-sophisticated policy gradient algorithms).

All coagent architectures used one fully-connected layer of $32$ coagents each for each document. That is, $32(5) = 160$ distinct coagents, and $32$ sets of unique coagent weights (since weights are shared across the five documents, see Figure \ref{fig:architecture}).
Each coagent uses binary (discrete) actions, and uses a linear basis (including a bias feature).
Each coagent uses softmax action selection with no temperature parameter.

Document selection (``Recommendation Generation'' in Figures \ref{fig:architecture} and \ref{fig:async_architecture}) is based on a simple ``vote'' across the network.
Specifically, an integer for each document is computed between $0$ and $32$ (from the $32$ coagents outputting binary actions), and the document with the greatest integer is that which is recommended.
Ties are broken by choosing the document at random from the documents that are tied.
Within the theoretical framework of \citet{kostas2020asynchronous}, this document selection portion of the network can be viewed as a coagent with no learnable parameters.

% For the bandit experiments, the hyperparameters used were a step-size of $0.0002457914452935442$ for the first layer, and $28.070098706460982$ for the second layer (an extensive hyperparameter search .

% For the non-bandit experiments, the hyperparameters used were:

\section{MSLR-WEB10K RL Baseline Plots}
\label{sec:baseline_mslr}

\begin{figure}
    \centering
    \includegraphics[width=1.0\linewidth]{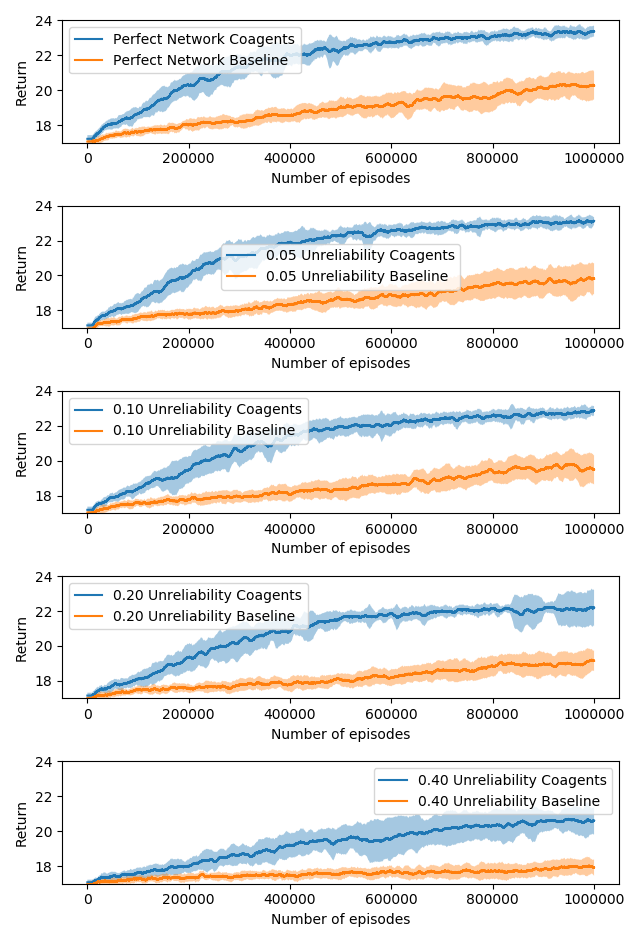}
    \caption{Baseline comparison for the MSLR-Web10K RL simulator}
    \label{fig:baseline_mslr_non_bandit}
\end{figure}

Figure \ref{fig:baseline_mslr_non_bandit} gives the baseline comparisons for the MSLR-WEB10K RL simulator.

\end{document}